\documentclass{article}
\usepackage{ijcai25}

\usepackage{times}
\usepackage{soul}
\usepackage{url}
\usepackage[hidelinks]{hyperref}
\usepackage[utf8]{inputenc}
\usepackage[small]{caption}
\usepackage{graphicx}
\usepackage{amsmath}
\usepackage{amsthm}
\usepackage{booktabs}
\usepackage{algorithm}
\usepackage{algorithmic}
\usepackage[switch]{lineno}

\usepackage{amssymb} 
\usepackage{multirow}
\usepackage{multicol}
\usepackage{booktabs}       % professional-quality tables
\usepackage{subfigure}
\title{Multimodal Regression for Enzyme Turnover Rates Prediction}

\author{
Bozhen Hu$^{1,2}$ \and
Cheng Tan$^1$ \and
Siyuan Li$^1$ \and
Jiangbin Zheng$^1$ \and
\\
Sizhe Qiu$^3$ \And
Jun Xia$^1$\thanks{Corresponding authors.} \and
Stan Z. Li$^1$\footnotemark[1]\\
\affiliations
$^1$AI Division, School of Engineering, Westlake University \\
$^2$Zhejiang University, 
$^3$Oxford University\\
\emails
\{hubozhen,tancheng,lisiyuan,zhengjiangbin,xiajun,stan.zq.li\}@westlake.edu.cn
}
\begin{document}

\maketitle

\begin{abstract}
    The enzyme turnover rate is a fundamental parameter in enzyme kinetics, reflecting the catalytic efficiency of enzymes. However, enzyme turnover rates remain scarce across most organisms due to the high cost and complexity of experimental measurements. To address this gap, we propose a multimodal framework for predicting the enzyme turnover rate by integrating enzyme sequences, substrate structures, and environmental factors. Our model combines a pre-trained language model and a convolutional neural network to extract features from protein sequences, while a graph neural network captures informative representations from substrate molecules. An attention mechanism is incorporated to enhance interactions between enzyme and substrate representations. Furthermore, we leverage symbolic regression via Kolmogorov-Arnold Networks to explicitly learn mathematical formulas that govern the enzyme turnover rate, enabling interpretable and accurate predictions. Extensive experiments demonstrate that our framework outperforms both traditional and state-of-the-art deep learning approaches. This work provides a robust tool for studying enzyme kinetics and holds promise for applications in enzyme engineering, biotechnology, and industrial biocatalysis.
\end{abstract}

\section{Introduction}
\label{Introduction}
\noindent
Enzymes, a type of protein found in cells, facilitate and accelerate chemical reactions crucial for various bodily functions, including muscle building, detoxification, and digestion. They often work in conjunction with other substances like stomach acid and bile~\cite{kosal2023digestion}, making the quantitative study of enzyme kinetics an important topic. The enzyme turnover rate (number) $k_\text{cat}$, which defines the maximum chemical conversion rate of a reaction, is a critical parameter for understanding the metabolism, proteome allocation, growth, and physiology of an organism~\cite{li2022deep}. Currently, the determination of enzyme kinetic parameters heavily relies on laboratory experimentation~\cite{nilsson2017metabolic}. These procedures are time-consuming, costly, and labor-intensive, resulting in a restricted database of experimentally determined kinetic parameters due to the absence of high-throughput techniques. Consequently, the repositories of $k_\text{cat}$ values in enzyme databases such as BRENDA~\cite{schomburg2017brenda} and SABIO-RK~\cite{wittig2018sabio} remain sparse when compared to the vast array of organisms and metabolic enzymes. While the sequence database UniProtKB~\cite{magrane2011uniprot} now boasts over 248M (million) protein sequences, the enzyme databases BRENDA and SABIO-RK contain only 17K (thousand) of experimentally measured kinetic parameters~\cite{yu2023unikp}. This scarcity of $k_\text{cat}$ data in databases underscores the need for the development of computational methods to predict $k_\text{cat}$.

\begin{figure}[thbp!]
\centering
\includegraphics[width=0.85\columnwidth]{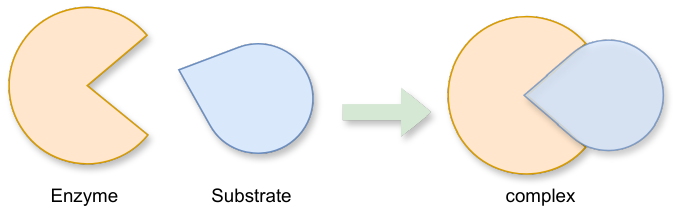} % Reduce the figure size so that it is slightly narrower than the column.
\caption{An illustration of the enzyme and substrate lock-and-key model.}
\label{fig-lock}
\end{figure}

With the rapid advancement of deep learning (DL) models, their applications have expanded to various fields, such as drug design, and enzyme reaction prediction~\cite{hu2024protgo}.
The prediction of kinetic parameters using DL-based techniques can be framed as a compound-protein interaction (CPI) prediction problem. This approach has been employed to predict various enzyme-related parameters, such as binding affinities ($k_\text{d}$), Michaelis–Menten constants ($k_\text{m}$), and enzyme turnover rates ($k_\text{cat}$). For example, DLKcat~\cite{li2022deep} is a famous DL approach for $k_\text{cat}$ prediction for metabolic enzymes from any organism using protein sequences and substrate structures. Based on this, the subsequent work is UniKP models~\cite{yu2023unikp}, which considers environmental factors such as pH and temperature that can impact enzyme kinetics significantly. Moreover, DLTKcat~\cite{qiu2024dltkcat} is proposed to quantify the influence of temperature on the $k_\text{cat}$ prediction, which demonstrates the feature importance of temperature under different cases~\cite{arroyo2022general}. However, these works do not explicitly learn relationships between $k_\text{cat}$ and environmental factors. Additionally, they employ different models to independently extract representations from enzyme sequences and substrates, lacking deeper interactions and associations. This is despite the profound connection between enzymes and substrates in enzymatic reactions, akin to the classic lock-and-key (or template) theory of enzyme specificity~\cite{prokop2012engineering}, as shown in Figure~\ref{fig-lock}. 

To address the limited availability of $k_\text{cat}$ values, we propose ProKcat, a multimodal deep learning framework that integrates enzyme sequences, substrate structures, and environmental factors. Enzyme sequences are processed using a pre-trained protein language model (LM)~\cite{lin2023evolutionary} and a convolutional neural network (CNN), while substrate molecules, represented in SMILES format~\cite{honda2019smiles}, are encoded with a graph neural network (GNN)~\cite{zhou2020graph,liuyue_HSAN}. An attention module enhances interactions between enzyme and substrate representations through soft alignment. To improve interpretability, we adopt Kolmogorov–Arnold Networks~\cite{liu2024kan} for symbolic regression, enabling explicit modeling of relationships among inputs such as temperature and $k_\text{cat}$. 

The contributions of this paper are summarized as follows:
(1) A comprehensive multimodal framework for predicting enzyme turnover rates, which fuses various types of information and generates meaningful representations.
(2) A specially designed interaction attention module that enhances the interactions derived from enzyme sequences and substrate structures.
(3) The first DL model to employ symbolic regression for $k_\text{cat}$ prediction with high efficiency, establishing clearer relationships between input factors and $k_\text{cat}$.
(4) Extensive experiments demonstrating that our proposed framework outperforms existing $k_\text{cat}$ prediction models, offering a promising approach for understanding enzyme kinetics with significant potential impact on biochemistry. 

\section{Related Works}
\subsection{Deep Learning-based Enzyme Turnover Rates Prediction}
In the initial stage, CNNs~\cite{lecun1995convolutional}, recurrent neural networks (RNNs)~\cite{grossberg2013recurrent}, and GNNs are employed to process enzyme and substrate features, followed by linear regression to predict enzyme kinetic parameters~\cite{lim2021review}. For instance, DLKcat~\cite{li2022deep} integrates a GNN for substrates and a CNN for proteins, facilitating high-throughput prediction of $k_\text{cat}$ and identifying critical amino acid residues influencing these predictions. Subsequently, EF-UniKP and Revised UniKP~\cite{yu2023unikp} are developed to predict temperature-dependent $k_\text{cat}$ values using an Extra Tree~\cite{sharaff2019extra} model, which processes concatenated representation vectors derived from a pre-trained LM, ProtT5~\cite{AhmedElnaggar2021ProtTransTC}, and a SMILES transformer model~\cite{honda2019smiles}. TurNuP~\cite{kroll2023turnover} characterizes chemical reactions through differential reaction fingerprints and represents enzymes using a modified Transformer model~\cite{vaswani2017attention}. GELKcat~\cite{du2023gelkcat} assigns weights to substrate and enzyme features using an adaptive gate network. In order to tackle the out-of-distribution problem, CatPred~\cite{boorla2024catpred} outputs predictions as gaussian distributions (including a mean and a variance). Recently, DLTKcat~\cite{qiu2024dltkcat} has shown potential in enzyme sequence design by predicting the impact of amino acid substitutions on $k_\text{cat}$ across various temperatures. However, these methods typically extract feature vectors from protein sequences and substrates independently, followed by a simple predictor, with limited interaction between the two modules. The evident relationships between temperature and $k_\text{cat}$ remain underexplored. To address these issues, we propose the incorporation of an additional attention module and the use of Kolmogorov-Arnold Networks (KANs) for symbolic regression of $k_\text{cat}$.

\subsection{Compound–Protein Interaction Prediction}
Traditional methods for CPI prediction typically involve screening candidates from a vast chemical space using various experimental assays~\cite{trott2010autodock} or employing molecular dynamics simulations~\cite{hollingsworth2018molecular}, both of which are inefficient. Recent advances in DL, however, have revolutionized CPI prediction by offering new approaches. Most DL-based techniques represent compounds as one-dimensional (1D) sequences or molecular graphs and proteins as 1D sequences, performing joint representation learning and interaction prediction within a unified framework. For instance, DeepConvDTI~\cite{lee2019deepconv} employs CNNs to extract low-dimensional representations of compounds and proteins, concatenates these representations, and then feeds them into fully connected (FC) layers to predict interactions. HyperattentionDTI~\cite{zhao2022hyperattentiondti} models complex non-covalent inter-molecular interactions among atoms and amino acids based on a CNN. More recently, PerceiverCPI~\cite{nguyen2023perceiver} has utilized a cross-attention mechanism to enhance the learning capabilities for compound and protein interaction representations. PSC-CPI~\cite{wu2024psc} captures the dependencies between protein sequences and structures through multi-scale contrastive learning.  

% DeepDTA~\cite{nguyen2021graphdta}, on the other hand, treats compounds as molecular graphs and leverages graph convolutional networks (GCNs)~\cite{kipf2016semi} instead of CNNs to learn compound representations. 

\begin{figure*}[tbp!]
\centering
\includegraphics[width=0.95\textwidth]{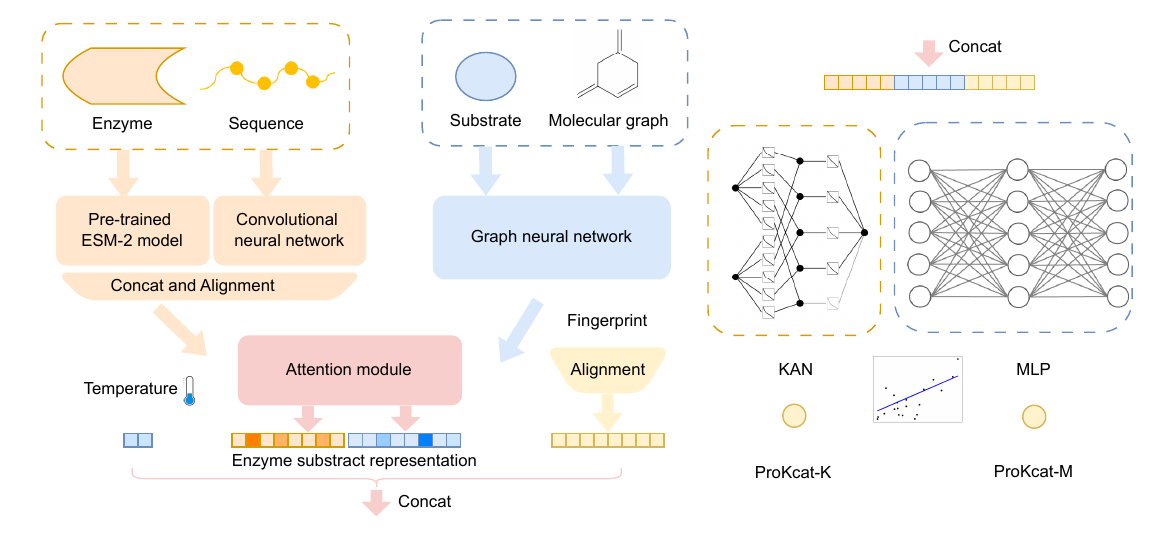} % Reduce the figure size so that it is slightly narrower than the column.
\caption{The overall framework of ProKcat integrates a pre-trained ESM-2 model and a CNN to analyze enzyme sequences, while a GNN processes substrate structures. To enhance the interaction between enzyme and substrate representations, an attention module is introduced. The attention-weighted outputs are then combined with molecular fingerprint features and temperature values. These fused features are fed into either a Kolmogorov–Arnold Network (ProKcat-K) or a multilayer perceptron (ProKcat-M) to predict $\log_{10}{k_\text{cat}}$ values.}
\label{fig-model}
\end{figure*}

\section{Methodology}
\subsection{Preliminaries}
\label{Preliminaries}
\paragraph{Problem Definitions and Notations.}
Let $\mathcal{S}=\{s_i\}_{i=1, \ldots, L_{p}}$ be a protein sequence with the length $L_p$, $s_i$ is the $i$-th amino acid. 
A SMILES format~\cite{honda2019smiles} of compound is transformed by RDKit~\cite{bento2020open} as a graph $G=(\mathcal{V}, \mathcal{E})$, where $\mathcal{V}=\{v_i\}_{i=1, \ldots, N_{v}}$ and $\mathcal{E}=\left\{\varepsilon_{i j}\right\}_{i, j=1, \ldots, N_{v}}$ denote the sets of vertices and edges with $N_{v}$ atoms, $\varepsilon_{i j}$ represents there is a chemical bond between atom $v_i$ and $v_j$. The environmental factor, temperature, is denoted as $T$.

Extended Connectivity Fingerprints (ECFPs)~\cite{rogers2010extended} are a type of topological fingerprint used to characterize molecular substructures. They describe the features of substructures by considering each atom and its surrounding circular neighborhoods within a specified diameter range, which has been demonstrated effective in small molecular characterization, similarity searching, and compound-protein representation learning~\cite{kroll2023turnover}. We use the open-source toolkit RDKit~\cite{bento2020open} in cheminformatics to calculate the fingerprint of compounds to obtain the feature vector $\mathbf{h}_f$ of length 1024 as another part of the input. Our purpose is for the regression of the $\log_{10}{k_\text{cat}} $ values.

\paragraph{Enzyme Turnover Rates.} Based on the Michaelis-Menten kinetics equation~\cite{michaelis1913kinetik}, $k_\text{cat}$ represents the number of substrate molecules converted to product per unit time by a single enzyme molecule when the enzyme is fully saturated with substrate, having units of $\text{time}^{-1}$, it is also called the enzyme turnover number.

\paragraph{Arrhenius Equation.} 
Several models have been proposed to elucidate the relationship between temperature and biological processes~\cite{brown2004toward}, with the Arrhenius equation emerging as the predominant choice in this field~\cite{arroyo2022general}. This equation describes how the rate constant of a chemical reaction varies with the absolute temperature according to the formula:
\begin{equation}
k=Ae^{\frac{-E}{k_BT}} 
\label{eq(2)}
\end{equation}
where $k$ represents a biological parameter (such as enzyme reaction rate), $k_B$ denotes Boltzmann’s constant, $T$ stands for the absolute temperature, $E$ signifies the effective activation energy for a specific process, and $A$ serves as a normalization constant that characterizes the process as a whole. It can be rearranged as $\ln_{}{k} =\frac{-E}{k_B}\frac{1}{T} +\ln_{}{A} $, meaning a linear relationship between $\ln_{}{k}$ and $\frac{1}{T}$.

\paragraph{Kolmogorov-Arnold Networks.} 
KAN~\cite{liu2024kan} is based on the Kolmogorov-Arnold representation theorem~\cite{tikhomirov1991representation,braun2009constructive}, which states that any multivariate continuous function $f:[0,1]^n \to \mathbb{R} $ on a bounded domain can be written as a finite composition of continuous functions of a single variable $x_i$ and the binary operation of addition,
\begin{equation}
f(x)=\sum_{j=1}^{2 n+1} \Phi_j\left(\sum_{i=1}^n \phi_{j, i}\left(x_i\right)\right)
\end{equation}
where $\phi_{j, i}:[0,1]^n \to \mathbb{R}$ and $ \Phi_j:\mathbb{R} \to \mathbb{R}$. This 2-layer width Kolmogorov-Arnold representation may not be smooth. Thus, \citeauthor{liu2024kan} extend this theorem by proposing a generalized architecture with wider and deeper KANs. Given a vector $\mathbf{x}\in\mathbb{R} ^{n_0}$, and  $\boldsymbol{\Phi}_l$ represents the $l$-th KAN layer, a KAN with $L_K$ layers can be expressed as
\begin{equation}
\operatorname{KAN}(\mathbf{x})=\boldsymbol{\Phi}_{L_{K-1}} \circ \cdots \circ \boldsymbol{\Phi}_1 \circ \boldsymbol{\Phi}_0 \circ \mathbf{x}
\label{eq(3)}
\end{equation}
where $\circ$ means function composition. The function $\phi(x)$ is composed of the sum of the SiLU basis function~\cite{paul2022sinlu} and a linear combination of B-splines during its implementation. B-splines are commonly employed for interpolating or approximating data points in a seamless fashion. The definition of a spline involves specifying its order, typically set at 3, and the number of intervals, refering to the number of segments or subintervals between adjacent control points~\cite{vaca2024kolmogorov}.

\subsection{Overall Framework}
The framework of ProKcat, as depicted in Figure~\ref{fig-model}, leverages multimodal features of enzymes and substrates. We use a currently prevalent pre-trained protein LM, ESM-2 (650M)~\cite{lin2022language} to generate protein sequence embeddings and adopt a CNN to learn from enzyme sequence on this task; then, the learned embeddings are concatenated. A feature alignment network with MLP layers standardizes latent features to a unified dimension $d$, yielding the feature vector of enzyme sequences $\mathbf{h}_p\in \mathbb{R}^{L_p\times d}$. Similarly, another feature alignment network transforms ECFPs feature vectors $\mathbf{h}_f\in \mathbb{R}^{1024}$ into the compound latent space $\mathbf{h}'_f\in \mathbb{R}^{d}$. For substrates represented as a graph $G=(\mathcal{V}, \mathcal{E})$, a graph attention network (GAT) is employed to learn from the compound graph structure. GNNs, known for their effectiveness in the enzyme kinetics parameters prediction tasks~\cite{li2022deep,qiu2024dltkcat}, update atom vectors and their neighboring vectors through neural network transformations. The GAT outputs real-valued molecular vector representations for substrates, denoted as $\mathbf{h}_c\in \mathbb{R}^{N_v \times d}$. 

\paragraph{Enzyme-Substrate Attention Module.}
In contrast to previous $k_\text{cat}$ prediction methods like  UniKP~\cite{yu2023unikp} and DLKcat~\cite{li2022deep}, our approach involves the design of an enzyme-substrate attention module to amplify interactions between enzymes and substrates. The enzyme vectors $\mathbf{h}_p\in \mathbb{R}^{L_p\times d}$ and compound vectors $\mathbf{h}_c\in \mathbb{R}^{N_v \times d}$ are derived in latent spaces. A soft alignment matrix $\mathcal{A}\in \mathbb{R}^{N_v\times L_p}$ is computed as follows:
\begin{equation}
\mathcal{A}=\sigma (\mathbf{h}_cW\mathbf{h}_p^\top   )
\end{equation}
$\mathcal{A}_{ij}$ represents the interaction strength between the $i$-th atom of compounds and $j$-th residue of proteins~\cite{li2022bacpi}.  The parameter matrix $W$ is trainable, and $\sigma(\cdot)$ denotes an activation function, such as the Tanh function~\cite{fan2000extended}, $^\top$ denotes transposition. Then, we compute the intermediate compound-to-protein and protein-to-compound features using FC layers and the soft alignment matrix $\mathcal{A}$,
\begin{equation}
    \begin{aligned}
        \mathbf{h}_{p2c}&=\mathcal{A}\cdot \text{FC}(\mathbf{h}_p)\\
 \mathbf{h}_{c2p}&=\mathcal{A}^\top\cdot \text{FC}(\mathbf{h}_c)
    \end{aligned}
    \label{eq(6)}
\end{equation}
where $\mathbf{h}_{p2c}\in \mathbb{R}^{N_v\times d}$ and $\mathbf{h}_{c2p}\in \mathbb{R}^{L_p\times d}$  represent the learned compound-to-protein and protein-to-compound features. The weights of protein features $\mathbf{h}_p$ should be derived from the compound-to-protein features $\mathbf{h}_{c2p}$ and the features $\mathbf{h}_p$ themselves, with similar operations for the weights of compound features. Therefore, the attention weights can be computed using the Softmax~\cite{joulin2017efficient} as 
\begin{equation}
    \begin{aligned}
        \mathbf{\alpha}_p &=\text{Softmax}(\text{FC}(\mathbf{h}_{c2p}\parallel \mathbf{h}_{p})) \\
 \mathbf{\alpha}_c &=\text{Softmax}(\text{FC}(\mathbf{h}_{p2c}\parallel \mathbf{h}_{c}))
    \end{aligned}
    \label{eq(7)}
\end{equation}
where $\mathbf{\alpha}_p\in \mathbb{R}^{L_p\times 1}$ and $\mathbf{\alpha}_c\in \mathbb{R}^{N_v \times 1}$ represent the normalized attention weights of protein and compound features, respectively, and $\parallel$ denotes the concatenation operation. Subsequently, the attention-weighted protein and compound features are computed through matrix multiplication as $\mathbf{h}'_{p} = \mathbf{\alpha}_p \cdot \mathbf{h}_{p}$ and $\mathbf{h}'_{c} = \mathbf{\alpha}_c \cdot \mathbf{h}_{c}$, resulting in the weighted features $\mathbf{h}'_{p} \in \mathbb{R}^{L_p \times d}$ and $\mathbf{h}'_{c} \in \mathbb{R}^{N_v \times d}$. 

This enzyme-substrate interaction process can be performed within a multi-head attention framework, where the distinct weighted protein and compound features from various heads are combined to derive the final enzyme-substrate representations. This mechanism is expressed as
\begin{equation}
    \begin{aligned}
        \mathbf{h}^{'multi}_p &= \parallel_{i}^{L_h} \text{Softmax}(\text{FC}^i(\mathbf{h}^i_{c2p}\parallel \mathbf{h}_{p}))\cdot \mathbf{h}_p \\
 \mathbf{h}^{'multi}_c &=\parallel_{i}^{L_h} \text{Softmax}(\text{FC}^i(\mathbf{h}^i_{p2c}\parallel \mathbf{h}_{c}))   \cdot \mathbf{h}_c
    \end{aligned}
    \label{eq(8)}
\end{equation}
where there are $L_h$ heads, $\mathbf{h}^{'multi}_p\in \mathbb{R}^{L_p \times L_hd}, \mathbf{h}^{'multi}_c\in \mathbb{R}^{N_v \times L_hd}$, then, linear projections are conducted to make the generated representations have a unified latent dimension.

\paragraph{Multivariable Regression.}
In latent spaces, we have acquired the weighted protein and compound representations $\mathbf{h}'_{p} \in \mathbb{R}^{L_p \times d}$ and $\mathbf{h}'_{c} \in \mathbb{R}^{N_v \times d}$, along with aligned fingerprint feature vectors $\mathbf{h}'_{f}\in \mathbb{R}^d$. A global average pooling operation is performed on these enzyme and substrate feature vectors to derive the comprehensive protein and compound level features, denoted as $\mathbf{h}''_{p} \in \mathbb{R}^{ d}$ and $\mathbf{h}''_{c} \in \mathbb{R}^{ d}$. Subsequently, these feature vectors are concatenated, resulting in $\mathbf{h}= [\mathbf{h}''_{p} \parallel \mathbf{h}''_{c} \parallel \mathbf{h}'_{f}]$, where $\mathbf{h} \in \mathbb{R}^{3d}$. Referring to the Arrhenius equation depicted in Eq.~\ref{eq(2)}, the environmental factor, temperature, $T$, along with its reciprocal $\frac{1}{T}$, are introduced as additional controllable variables within the latent space, which are characterized by the learned features $\mathbf{h} \in \mathbb{R}^{3d}$. 

In this paper, we are presented with a choice between two approaches: one entails the utilization of MLPs for regression to predict the $\log_{10}{k_\text{cat}}$ values, designated as the ProKcat-M; while the alternative option involves constructing a KAN for the prediction of these values, known as the ProKcat-K. Referring to Eq.~\ref{eq(3)}, the KAN module with $L_K$ layers in the ProKcat-K model can be mathematically formulated as
\begin{equation}
\log_{10}{k_\text{cat}} =\boldsymbol{\Phi}_{L_K-1} \circ \cdots \circ \boldsymbol{\Phi}_1 \circ \boldsymbol{\Phi}_0 \circ (\mathbf{h} \parallel T \parallel \frac{1}{T})
\label{(9)}
\end{equation}
here, the input feature vectors comprise two components: one being the learned enzyme-substrate weighted representations derived from a deep neural network, and the other component consisting of the environmental variables, specifically temperature, $T$ and $ \frac{1}{T}$. The former component is acquired from a black-box model, implying the inability to explicitly define a formula expressing the representations and their associations with the input enzyme sequences and substrate compounds. Nevertheless, a clear linear relationship exists between $\log_{10}{k_\text{cat}}$ and $\frac{1}{T}$ based on the Arrhenius equation, indicating that the input variables $T$ and $\frac{1}{T}$ exhibit explicit relationships with the output $\log_{10}{k_\text{cat}}$.

Conventional symbolic regression techniques are commonly intricate and arduous to diagnose. Their outcomes frequently lack clear intermediate insights. In contrast, KANs engage in continuous exploration, yielding smoother and more robust results, which have been demonstrated superior performance in function representation compared to MLPs across various tasks such as regression and partial differential equation (PDE) solving~\cite{liu2024kan}. As a result, the ProKcat-K model can establish direct associations between input variables and the output, offering partial interpretability. Notably, with the given pre-trained sequence-substrate representations and temperature inputs, the direct calculation of $\log_{10}{k_\text{cat}}$ based on the learned formula becomes feasible. This capability holds significant implications for the fields of bioinformatics and biochemistry.

\begin{figure}[tbp]
\centering
\includegraphics[width=0.7\columnwidth]{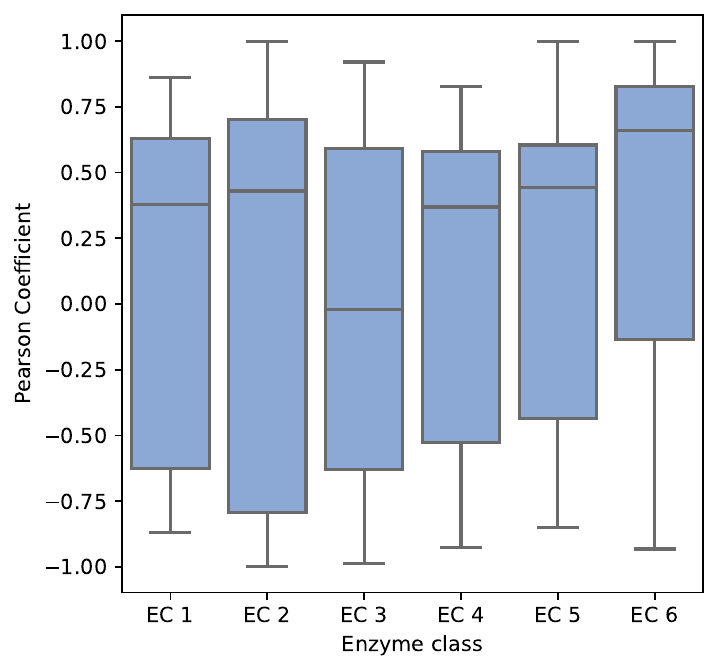} % Reduce the figure size so that it is slightly narrower than the column.
\caption{Pearson correlations between temperature and $k_\text{cat}$ (with a significance level of $p < 0.05$) across diverse enzyme classes.}
\label{Coefficient}
\end{figure}

% \begin{figure}[tbp!]
% \centering
% % \subfigure[Count of $k_\text{cat}$]{\includegraphics[width=0.35\textwidth]{Figs/median_kcat.pdf}
% % \label{median_kcat}}
% % \quad
% \subfigure[Count of temperature]{\includegraphics[width=0.53\columnwidth]{Figs/temperature.pdf}\label{temperature}}
% % \quad
% \subfigure[Pearson correlations]{\includegraphics[width=0.43\columnwidth]{Figs/Coefficient.pdf}\label{Coefficient}}
% % \quad
% \caption{Data statistics. (a) The histogram displays the distribution of temperature values in Celsius, indicating a median optimal temperature of 37 $^\circ\mathrm {C} $. (b) Pearson correlations between temperature and $k_\text{cat}$ (with a significance level of $p < 0.05$) across diverse enzyme classes, the median values are denoted by the horizontal line within the box plots.  }
% \label{fig-datasets}
% % \vspace{-10pt}
% \end{figure}

\section{Experiments}
\subsection{Experimental Setup}
\paragraph{Datasets.}
The fundamental details pertaining to enzymatic reactions are sourced from the databases BRENDA~\cite{schomburg2017brenda} and SABIO-RK~\cite{wittig2018sabio}, encompassing key attributes such as the enzyme commission (EC) number, enzyme type, operating temperature, $k_\text{cat}$ values, substrate name, Uniprot ID, etc. Leveraging the UniProt ID and substrate designation, the enzyme sequences and substrate SMILES strings are retrieved from the UniProtKB repository~\cite{magrane2011uniprot} and the PubChem compound database~\cite{kim2023pubchem}. The process mirrors the procedures adopted by other DL-based models for $k_\text{cat}$ prediction~\cite{li2022deep,qiu2024dltkcat}. Following the operations in DLKcat~\cite{li2022deep}, redundant entries sharing identical SMILE strings, amino acid sequences, operational temperatures, and $k_\text{cat}$ values are eliminated from the raw dataset, retaining solely the highest $k_\text{cat}$ value when other attributes are equivalent. The resulting dataset comprises over 10,000 entries from BRENDA and 4,000 entries from SABIO-RK. To address the uneven distribution of temperature values, an oversampling technique is employed to augment the dataset by duplicating entries at lower and higher temperature ranges~\cite{qiu2024dltkcat}.

% Upon analyzing the resultant dataset, a predominant portion of the entries originate from the BRENDA database. The distribution of optimal temperature values is illustrated in Figure~\ref{temperature}, revealing a median optimal temperature of 37 $^\circ\mathrm {C}$ for the enzymes. This observation suggests a bias towards mammalian and microorganism data associated with mammalian habitats, particularly emphasizing temperatures ranging between 20 and 40 $^\circ\mathrm {C}$. To address the uneven distribution of temperature values, an oversampling technique is employed to augment the dataset by duplicating entries at lower and higher temperature ranges~\cite{qiu2024dltkcat}.

Enzymes exhibit sensitivity to temperature variations. A gradual increment in temperature typically enhances enzymatic reaction rates; however, beyond a certain threshold, further temperature values lead to a decline in reaction speed. This phenomenon arises due to temperature's dual impact on enzymatic reactions, while elevated temperatures can expedite reaction kinetics, they also accelerate enzyme denaturation, thereby diminishing active enzyme concentrations and catalytic efficiency~\cite{arcus2020temperature}. Considering the intricate interplay between temperature and $k_\text{cat}$ in biological reactions, the analysis of Pearson correlations between temperature and $k_\text{cat}$ (p-value $<$ 0.05) across six enzyme classes, encompassing top-level EC numbers 1-6 including oxidoreductase, transferase, hydrolase, lyases, isomerases, and ligases, is conducted. The correlation analysis, depicted in Figure~\ref{Coefficient}, reveals a positive relationship between temperature and $k_\text{cat}$ for all enzyme classes except hydrolases, as indicated by the median values.

Following previous works UniKP~\cite{yu2023unikp} and DLTKcat~\cite{qiu2024dltkcat},  the resulted dataset is partitioned into training and testing subsets at a ratio of 90\% and 10\%, with the training set further segmented into a validation subset comprising one-tenth of the training data. Other experimental details are presented in the appendix.

\paragraph{Baselines.} The baselines selected for comparison with our proposed model in predicting $k_\text{cat}$ can be categorized into two primary groups. The first group comprises machine learning (ML) models, such as Linear Regression~\cite{gross2003linear}, Decision Tree~\cite{song2015decision}, AdaBoost~\cite{hastie2009multi}, Support Vector Regressor~\cite{awad2015support}. The second group consists of DL-based approaches, including CNN~\cite{lecun1995convolutional}, RNN~\cite{grossberg2013recurrent}, MLP regressor~\cite{dutt2022multilayer}, these three models are the basic DL networks. Additionally, the specific $k_\text{cat}$ prediction methods based on DL techniques are considered, including DLKcat~\cite{li2022deep}, UniKP~\cite{yu2023unikp}, TurNuP~\cite{kroll2023turnover}, GELKcat~\cite{du2023gelkcat}, and DLTKcat~\cite{qiu2024dltkcat}.  Five different initializations are conducted to evaluate these methods. The mean and standard deviation values are reported.

% \begin{figure}[htbp]
% \centering
% \includegraphics[width=0.6\columnwidth]{Figs/gaussian.pdf} % Reduce the figure size so that it is slightly narrower than the column.
% \caption{The test performance of ProKcat-M without the attention module.}
% \label{gaussian}
% \end{figure}

\begin{table*}[tp]
  \centering
  \begin{tabular}{llcccc}
    \toprule
    Category & Method & RMSE$\downarrow  $ & PCC$\uparrow $ & MAE$\downarrow  $ & $R^2$$\uparrow $\\
    \midrule
    \multirow{4}{*}{ML-based}  & Linear Regression~\cite{gross2003linear} & $1.18_{ \pm 0.05}$ & $0.64_{ \pm 0.02}$ & $0.88_{ \pm 0.06}$ & $0.38_{ \pm 0.02}$ \\
    & Support Vector~\cite{awad2015support} & $1.35_{ \pm 0.05}$ & $0.44_{ \pm 0.03}$ & $1.04_{ \pm 0.09}$ & $0.19_{ \pm 0.03}$ \\
     & Decision Tree~\cite{song2015decision} & $1.27_{ \pm 0.07}$ & $0.65_{ \pm 0.03}$ & $0.83_{ \pm 0.01}$ & $0.29_{ \pm 0.01}$ \\
     & AdaBoost~\cite{hastie2009multi} & $1.34_{ \pm 0.01}$ & $0.48_{ \pm 0.02}$ & $1.07_{ \pm 0.01}$ & $0.21_{ \pm 0.02}$ \\
     \hline
     \multirow{3}{*}{DL-based} & CNN~\cite{lecun1995convolutional} & $1.42_{ \pm 0.16}$ & $0.34_{ \pm 0.06}$ & $1.11_{ \pm 0.05}$ & $0.10_{ \pm 0.05}$ \\
     & RNN~\cite{grossberg2013recurrent} & $1.35_{ \pm 0.12}$ & $0.44_{ \pm 0.05}$ & $1.05_{ \pm 0.06}$ & $0.19_{ \pm 0.10}$ \\
     & MLP regressor~\cite{dutt2022multilayer} & $1.08_{ \pm 0.08}$ & $0.72_{ \pm 0.05}$ & $0.81_{ \pm 0.04}$ & $0.48_{ \pm 0.04}$ \\
     \hline
     \multirow{6}{*}{$k_\text{cat}$ Prediction Models}
    & DLKcat~\cite{li2022deep} & $1.13_{ \pm 0.15}$ & $0.75_{ \pm 0.06}$ & $0.73_{ \pm 0.08}$ & $0.47_{ \pm 0.11}$ \\
    & UniKP~\cite{yu2023unikp} & $0.82_{ \pm 0.01}$ & $0.85_{ \pm 0.02}$ & $0.58_{ \pm 0.01}$ & $0.67_{ \pm 0.02}$ \\
    & TurNuP~\cite{kroll2023turnover} & $0.89_{ \pm 0.01}$ & $0.62_{ \pm 0.04}$ & $0.91_{ \pm 0.03}$ & $0.38_{ \pm 0.04}$ \\
    & GELKcat~\cite{du2023gelkcat} & $1.00_{ \pm 0.04}$ & $0.78_{ \pm 0.03}$ & $0.69_{ \pm 0.05}$ & $0.58_{ \pm 0.03}$ \\
    & DLTKcat~\cite{qiu2024dltkcat} & $0.91_{ \pm 0.02}$ & $0.80_{ \pm 0.02}$ & $0.63_{ \pm 0.03}$ & $0.65_{ \pm 0.02}$
    \\
    \cmidrule(lr){2-6}
    & ProKcat-M (Proposed) & $\textbf{0.71}_{\pm 0.01}$ & $\textbf{0.88}_{\pm 0.02}$ & $\textbf{0.48}_{\pm 0.02}$ & $\textbf{0.74}_{\pm 0.01}$ \\
    \bottomrule
  \end{tabular}
    \caption{Performance comparison of different models. The best results are shown in bold.}
   \label{kcat-table}
\end{table*}

\paragraph{Metrics.} To evaluate the efficacy of our proposed model, a comprehensive set of metrics is employed, encompassing the coefficient of determination ($R^2$), the Pearson correlation coefficient (PCC), the root mean square error (RMSE), and the mean absolute error (MAE)~\cite{yu2023unikp}. Eq. 10 provided in the appendix elucidates that $R^2$ signifies the proportion of variance explained, with values ranging between 0 and 1, where a value of 1 denotes a perfect model fit. The PCC quantifies the linear relationship between predicted and actual values, varying from -1 to 1, where 1 indicates a perfect positive linear correlation, -1 represents a perfect negative linear correlation, and 0 signifies no linear association. RMSE serves as a metric to assess the disparities between predicted and observed values. MAE offers an alternative measure of the deviations between predicted and actual outcomes.

\subsection{Results of Enzyme Turnover Rates Prediction}
Table~\ref{kcat-table} presents a comparative analysis of model performance in predicting $k_{\text{cat}}$ values. The results indicate that ML-based approaches demonstrate competitive performance in terms of RMSE, PCC, MAE, and $R^2$ when compared to DL-based models, such as CNN and RNN. The limitation observed in the DL models is attributed to their complex network architecture requirements and the challenge posed by the relatively small dataset size, hindering the complete training of DL models. These DL-based methods also have higher standard deviation values.

In contrast, models like UniKP, TurNup, DLTKcat, and our proposed ProKcat-M leverage large-scale pre-trained models' embeddings, leading to improved predictive performance. For instance, UniKP utilizes ProtT5-XL for encoding enzyme sequences and employs a pre-trained LM, SMILES Transformer model, to represent substrate structures, resulting in the second-highest ranking across all performance metrics. Our proposed model, ProKcat-M, incorporates a state-of-the-art protein sequence encoder and features an enzyme-substrate attention module, surpassing all other models with the lowest RMSE, MAE, highest PCC, and $R^2$ values. These results underscore the superior predictive capacity of ProKcat-M in estimating $k_{\text{cat}}$ values. 

\begin{table}
  \centering
  \begin{tabular}{lcccc}
    \toprule
    Method & RMSE$\downarrow  $ & PCC$\uparrow $ & MAE$\downarrow  $ & $R^2 \uparrow$ \\
    \midrule
    ProKcat-M & 0.71 & 0.88 & 0.48 & 0.74 \\
    \hline
    w/o attention & 0.89 & 0.82 & 0.56 & 0.67 \\
    w/o CNN & 1.04 & 0.77 & 0.70 & 0.56 \\
    w/o ESM-2 & 0.90 & 0.81 & 0.65 & 0.66 \\
    w/o enzyme & 1.26 & 0.59 & 0.96 & 0.32  \\
    w/o substrate & 1.17 & 0.65 & 0.85 & 0.41 \\
    w/o fingerprint & 1.11 & 0.68 & 0.78 & 0.47 \\

    \bottomrule
  \end{tabular}
  \caption{Ablation of ProKcat-M, we compare it with the models removing the attention module (w/o attention) and the models removing the CNN, ESM-2 embeddings, enzyme sequences (CNN and ESM-2 embeddings), substrate structures (GNN), the fingerprint feature vectors ($\mathbf{h}_f$).}
  \label{table-ablation}
\end{table}

\paragraph{Ablation Study.} 
Table~\ref{table-ablation} presents the results of the ablation study on ProKcat-M, comparing it with various models where specific components are removed. ProKcat-M, the full model, achieves promising performance, the subsequent rows in the table represent the performance of models with specific components removed:
\begin{itemize}
    \item[-]  Removing the attention module (w/o attention) leads to a moderate decrease in performance across all metrics.
    \item[-] Omitting CNN or the ESM-2 embeddings results in a higher RMSE and MAE, lower PCC and $R^2$, indicating the importance of CNN and EMS-2 embeddings in the model. Both CNN and EMS-2 embeddings are extracted from enzyme sequences, the former is learned on this task, but the latter is trained on large-scale protein sequences in an unsupervised way.  
    \item[-] Removing enzyme sequences (w/o enzyme) significantly deteriorates the model's predictive ability, as evidenced by the substantial increase in RMSE and MAE.
    \item[-] Excluding substrate structures (w/o substrate) and fingerprint feature vectors (w/o fingerprint) also lead to decreased performance.
\end{itemize}

\begin{figure}[tbp]
\centering
\includegraphics[width=0.7\columnwidth]{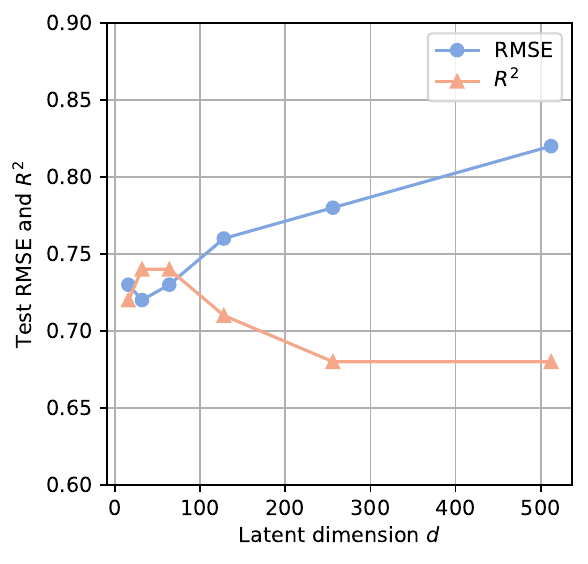} % Reduce the figure size so that it is slightly narrower than the column.
\caption{The test performance of ProKcat-M when choosing different latent dimensions $d$.}
\label{dimension}
\end{figure}

The latent dimension, denoted as $d$, serves as a critical network hyperparameter. Leveraging the AutoML toolkit NNI~\cite{nni2021}, we conducted a search for the optimal value with search space $\{ 16, 32, 64, 128, 256 \}$. The results are shown in Figure~\ref{dimension}, determining that a lower value of $d=32$ surpasses larger alternatives.

% This finding may suggest that the training dataset is relatively limited for this task, and a modest value of $d$ is deemed appropriate to achieve satisfactory outcomes. 

\subsection{Results of Symbolic Regression}
As depicted in Eq.~\ref{(9)}, symbolic regression is performed utilizing a KAN model, where the input to the KAN comprises the concatenated representations $\mathbf{h} \in \mathbb{R}^{3d}$, temperature $T$, and $\frac{1}{T}$. On this task, many KAN models appear to suffer from overfitting, there exists a substantial disparity between the training and testing metrics. For instance, a 5-depth KAN model with 28K parameters achieves an RMSE of 0.66 on the training set but 0.99 on the test set. Adjustments such as reducing network depth or training duration have been demonstrated to enhance KAN performance. KANs are renowned for their proficiency in regression or PDE solving in mathematical and physical domains. Consequently, addressing the challenge of overfitting becomes crucial in tackling this intricate, high-dimensional problem using KANs. This outcome underscores the effectiveness of KANs in achieving competitive performance with a relatively modest parameter count. For example, a 5-depth KAN model with 28K parameters attains an $R^2$ of 0.41, whereas a 3-depth KAN model with 25K parameters achieves an $R^2$ of 0.50.

Beyond mitigating the issue of overfitting in KANs, our main aim is to derive a comprehensive formula to establish the relationships between input and output variables. Given that the embeddings $\mathbf{h}$ originate from a deep neural network, our focus centers on explicitly modeling the correlations between $T, \ \frac{1}{T}$, and $k_\text{cat}$. To derive an explicit and concise equation, we perform a linear projection on the feature vectors $\mathbf{h}''_{p}: \mathbb{R}^{d} \to \mathbb{R}^{1}, \mathbf{h}''_{c}: \mathbb{R}^{d} \to \mathbb{R}^{1}, \mathbf{h}'_{f}: \mathbb{R}^{d} \to \mathbb{R}^{1}$, resulting in the concatenated feature vectors $\mathbf{h}: \mathbb{R}^{3d} \to \mathbb{R}^{3}$. 

\begin{figure}[tbp]
\centering
\includegraphics[width=0.7\columnwidth]{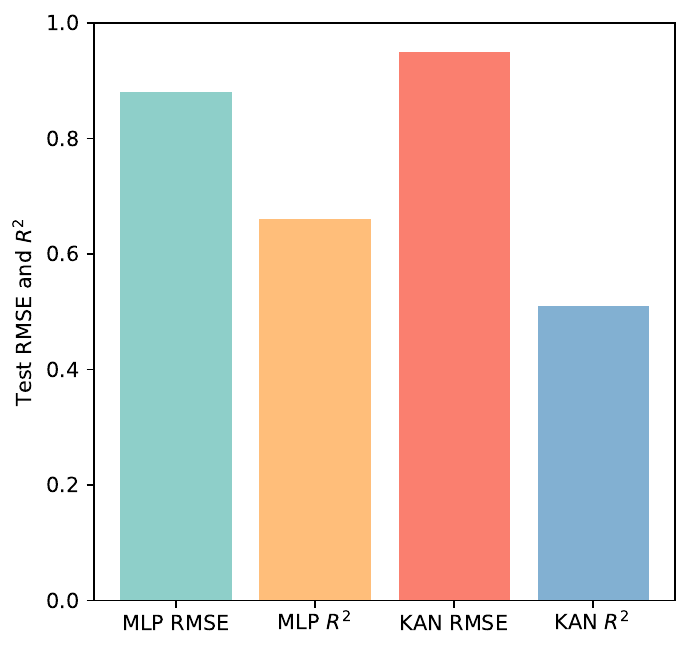} % Reduce the figure size so that it is slightly narrower than the column.
\caption{Comparisons of ProKcat-M and ProKcat-K on the same test set in terms of RMSE and $R^2$.}
\label{MLP-kan-test}
\end{figure}

In our experiments, we observe that by reducing the dimensionality of the combined vector ($\mathbf{h}$) to 3, a streamlined 2-layer KAN model can be developed. Specifically, utilizing B-spline details with 5 intervals, order 3, and steps 5. To ensure a fair comparison between KAN and MLP, we implement a 2-layer MLP with the latent dimension ranging from $3d+2$ to 1. Both the 2-layer KAN model and the 2-layer MLP model possess nearly equivalent trainable parameters, approximately 0.1K. From Figure~\ref{MLP-kan-test}, we observe that the 2-layer KAN model yields results comparable to a 2-layer MLP. Notably, the 2-layer KAN model demonstrates higher efficiency, with an inference time of 1 ms (millisecond) per sample, in contrast to 3.6 ms per sample for the 2-layer MLP.

% \section{Regression Function}
% \label{Regression_Functions}

% \begin{figure}[htbp]
% \centering
% \includegraphics[width=0.9\columnwidth]{Figs/kan-formula.png} % Reduce the figure size so that it is slightly narrower than the column.
% \caption{Visualization of the learned KAN model.}
% \label{fig-kan-model}
% \end{figure}

% \begin{figure*}[hbp!]
% \centering
% \includegraphics[width=0.6\textwidth]{Figs/kan-formula.png} % Reduce the figure size so that it is slightly narrower than the column.
% \caption{Visualization of the learned KAN model.}
% \label{fig-kan-model}
% \end{figure*}

According to the Arrhenius equation (Eq.~\ref{eq(2)}), a significant linear correlation is observed between the natural logarithm of the rate constant ($\ln{k}$) and the reciprocal of temperature ($\frac{1}{T}$), indicating a predominantly empirical relationship. It is noted that in practical scenarios, the relationships are more complicated with data noise existed. The symbolic regression-derived function of the 2-layer KAN model (ProKcat-K) is expressed as:

\begin{equation}
\begin{aligned}
    & 0.02*|9.92*\mathbf{h}''_{p} - 1.29| - 0.09*|3.57*\mathbf{h}''_{c} - 0.21| + \\ &0.01*e^{3.57*\mathbf{h}'_{f}} - 0.03*|9.22*\frac{1}{T} - 3.71| - \\ & 0.02*e^{-9.22*T} + 0.21
\end{aligned}
\label{eq(11)}
\end{equation}
The equation reveals a linear association between $\frac{1}{T}$ and $\log_{10}k_\text{cat}$, aligning with the principles of Eq.~\ref{eq(2)}. This underscores the reliability of our learned regression function to a certain degree. When given the pre-trained representations and temperature inputs, the direct calculation of $\log_{10}{k_\text{cat}}$ can be achieved. It is the first time to derive such an equation in the DL-based $k_\text{cat}$ prediction field, which is promising.

\section{Conclusion}
\label{Conclusion}
This paper presents a novel multimodal framework that integrates enzyme sequences, substrate compound structures, and additional features using a pre-trained language model LM, a convolutional neural network, and a graph neural network to predict $k_\text{cat}$ values. The proposed enzyme-substrate attention module effectively learns attention weights by capturing relationships between sequence and atomic-level features. Recognizing the critical importance of enzyme–compound interactions, the ProKcat-M model achieves superior predictive performance compared to existing baselines. Furthermore, the ProKcat-K variant employs the Kolmogorov–Arnold Network architecture to perform accurate $k_\text{cat}$ predictions with lower inference time, while also yielding an explicit equation that relates input variables to the output. However, a key limitation of the KAN model is its sensitivity to overfitting, which necessitates careful architectural and regularization design.

\section*{Acknowledgments}
This work was supported by the National Natural Science Foundation of China (Project No. 624B2115, 623B2086 and U21A20427), the Science \& Technology Innovation 2030 Major Program (Project No. 2021ZD0150100), the Center of Synthetic Biology and Integrated Bioengineering at Westlake University (Project No. WU2022A009), the Westlake University Industries of the Future Research Program (Project No. WU2023C019), and the Zhejiang Key Laboratory of Low-Carbon Intelligent Synthetic Biology (Project No. 2024ZY01025).

\section*{Contribution Statement}
Bozhen Hu and Cheng Tan contributed equally to this work. They jointly designed the model, conducted the experiments, and co-authored the manuscript. Siyuan Li and Jiangbin Zheng provided conceptual guidance. Jun Xia and Stan Z. Li reviewed and revised the manuscript. We also thank Sizhe Qiu, a Ph.D. student in the Department of Engineering Science at the University of Oxford, for his valuable insights and biological expertise.

\bibliographystyle{named}
\bibliography{ijcai25}

\end{document}